\documentclass[letterpaper]{article} 
\usepackage{aaai2027}  
\nocopyright
\usepackage[hyphens]{url}  
\usepackage{graphicx} 
\usepackage{natbib}  
\usepackage{caption} 
\usepackage{algorithm}
\usepackage{algorithmic}

\usepackage{booktabs}
\usepackage{multirow}

\usepackage{amsmath}
\usepackage{amssymb}
\usepackage{mathtools}
\usepackage{xspace}
\usepackage{xurl}

\newcommand{\ours}{GUI-Lens\xspace}

\title{GUI-Lens: Enabling Precise GUI Grounding for Off-the-Shelf VLMs via Coarse-to-Fine Cropping}

\author{
Zichuan Fu\textsuperscript{1}\thanks{Equal contribution.}\textsuperscript{,}\thanks{Work done during an internship at Tencent.},
Shirong Wang\textsuperscript{1}\footnotemark[1],
Wenlin Zhang\textsuperscript{1},
Guojing Li\textsuperscript{1},
Yimin Deng\textsuperscript{1},
Jingtong Gao\textsuperscript{1},\\
Junjia Qi\textsuperscript{1},
Hanyu Yan\textsuperscript{1},
Yefeng Zheng\textsuperscript{3},
Xiaopeng Li\textsuperscript{1},
Wanyu Wang\textsuperscript{1},
Xian Wu\textsuperscript{2},
Xiangyu Zhao\textsuperscript{1}\thanks{Corresponding author.}
}

\affiliations{
\textsuperscript{1}City University of Hong Kong,
\textsuperscript{2}Tencent Jarvis Lab,
\textsuperscript{3}Westlake University\\
\texttt{zc.fu@my.cityu.edu.hk, shirowang6-c@cityu.edu.hk, xianzhao@cityu.edu.hk}
}

\begin{document}

\maketitle

\begin{abstract}
GUI grounding maps natural-language instructions to click locations and is essential for reliable GUI agents.
The task remains difficult on high-resolution, densely populated interfaces because a vision-language model (VLM) may recognize a requested control without locating it precisely enough for interaction.
Most existing methods provide various forms of localization assistance, but still rely on a direct click prediction, allowing visual ambiguity or an inaccurate initial estimate to propagate to the final result.
In this paper, we introduce \ours, a coarse-to-fine grounding framework that allows a general-purpose VLM to determine the target through active visual observations.
Specifically, \ours extracts OCR text and detected UI components from the screenshot and presents their positions as coordinate references.
Using the instruction, the current view, and these references, the VLM selects the region and scale of the next view, which is cropped and enlarged to provide finer visual details.
This process continues over successively focused views until the target is determined.
Proposed crops and clicks are checked against the instruction throughout the process, and the final local position is mapped back to the original screen coordinates.
Experiments on five GUI grounding benchmarks and six general-purpose base models show that \ours consistently improves grounding accuracy, by up to $9.3\times$ on ScreenSpot-Pro.
\begin{links}
    \link{Code}{https://github.com/wangshirong123/GUI-Lens}
\end{links}
\end{abstract}


\begin{figure}[t]
    \centering
    \includegraphics[width=\columnwidth]{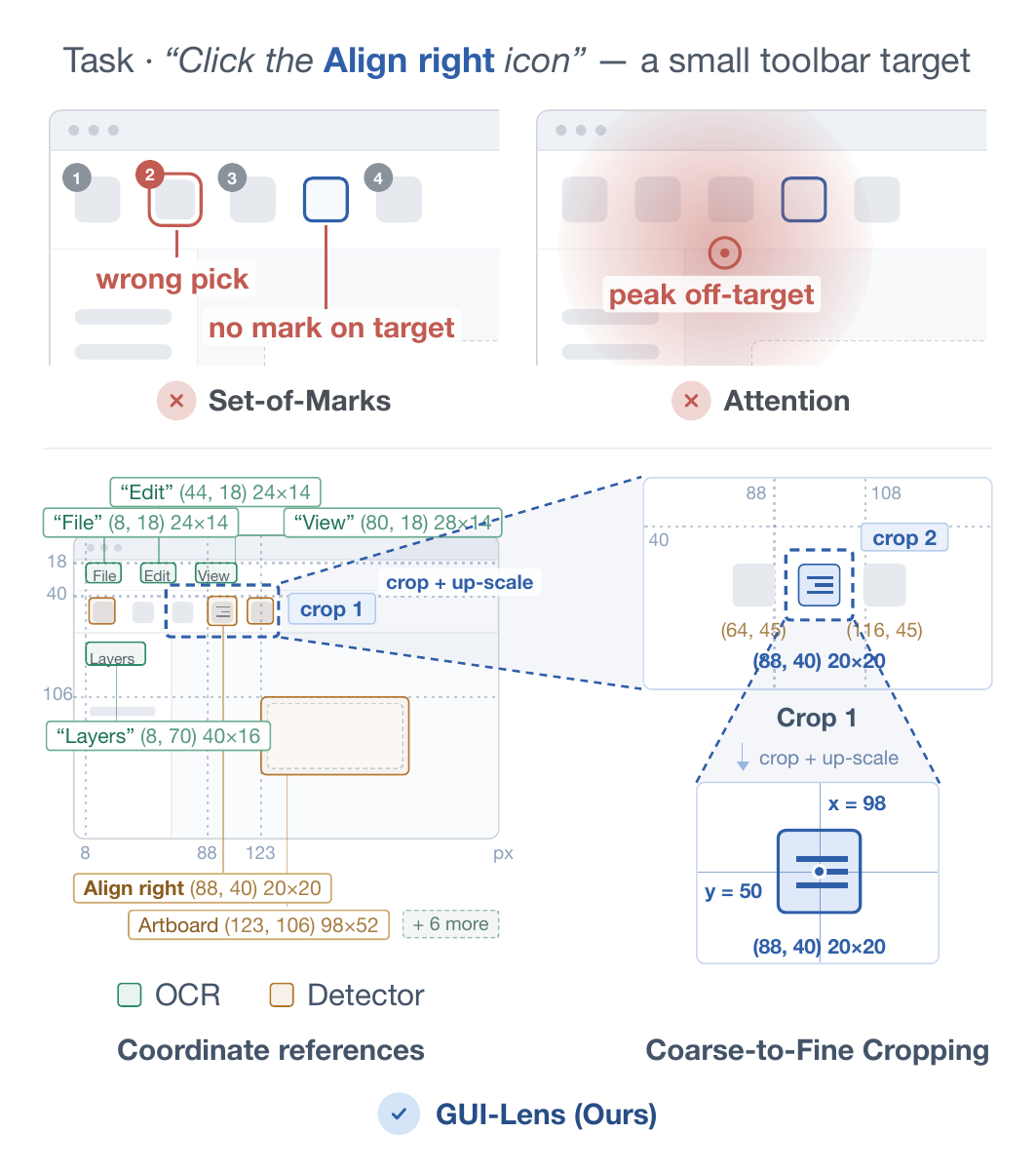}
    \caption{Comparison of GUI grounding methods on a small toolbar target. \ours uses coordinate references and coarse-to-fine cropping to achieve precise localization.}
    \label{fig:gui_grounding_motivation}
\end{figure}

\section{Introduction}
\label{sec:intro}

GUI agents aim to complete digital tasks by operating web pages, mobile applications, desktop systems, and professional software through their graphical interfaces~\citep{appagent,osworld,guisurvey,acusurvey}.
Recent vision-language models (VLMs) have accelerated this direction by jointly interpreting screenshots, natural-language instructions, and interface states, substantially improving task understanding and action planning~\citep{cogagent,aguvis,uitars}.
For actions that require direct interaction with on-screen elements, a complete GUI agent must determine the precise click location of the target control based on the instruction.
This capability, known as GUI grounding, is essential for converting action plans into reliable interface interactions in real applications.

However, reliable GUI grounding remains difficult because understanding the target specified by an instruction is not equivalent to determining its precise click location in the original screen coordinate system.
On high-resolution, densely populated interfaces, targets are often small and visually similar to neighboring elements~\citep{screenspotpro,mmbench_gui,ui_vision}.
Resizing the input image can further remove the local details needed to distinguish them.
Even when a VLM correctly recognizes the target, a small localization error may place the click on an adjacent control.

Existing GUI grounding methods commonly provide spatial cues through detected components or model-internal attention.
Component-based approaches expose detected or indexed regions as selectable candidates, but they cannot select a target that is missing from the candidate set~\citep{setofmark}.
As shown in Figure~\ref{fig:gui_grounding_motivation}, Set-of-Marks omits the requested toolbar control.
Attention-based approaches derive target locations from internal model responses, but the strongest attention response may not correspond to the intended clickable element~\citep{gui_aima}.
In Figure~\ref{fig:gui_grounding_motivation}, the attention response falls outside the target icon.
These limitations raise a question: can multi-step visual reasoning localize a GUI target at progressively finer granularity before the VLM predicts its final coordinates?

\begin{figure}[t]
    \centering
    \includegraphics[width=\linewidth]{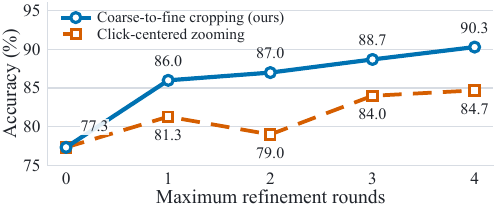}
    \caption{Comparison between coarse-to-fine cropping and click-centered zooming on ScreenSpot-Pro with GPT-5.5.}
    \label{fig:crop_vs_zoom_rounds}
\end{figure}

To address this challenge, zoom-based refinement methods have explored multi-step localization by repeatedly enlarging the region around a predicted click location~\citep{zoomclick}.
However, when each new view is centered on the preceding click prediction, an early localization error can constrain subsequent visual observations.
We instead draw inspiration from active visual attention, which adaptively selects image regions for high-resolution processing~\citep{mnih2014attention}.
Following this principle, each crop is treated as a new visual observation whose region and scale are selected by the VLM.
We compare coarse-to-fine cropping with click-centered zooming under the same evaluation setting.
As shown in Figure~\ref{fig:crop_vs_zoom_rounds}, coarse-to-fine cropping achieves higher accuracy under every nonzero refinement setting, showing that model-selected regions and scales are more effective than click-centered views for multi-step localization.

\begin{figure*}[t]
    \centering
    \includegraphics[width=\linewidth]{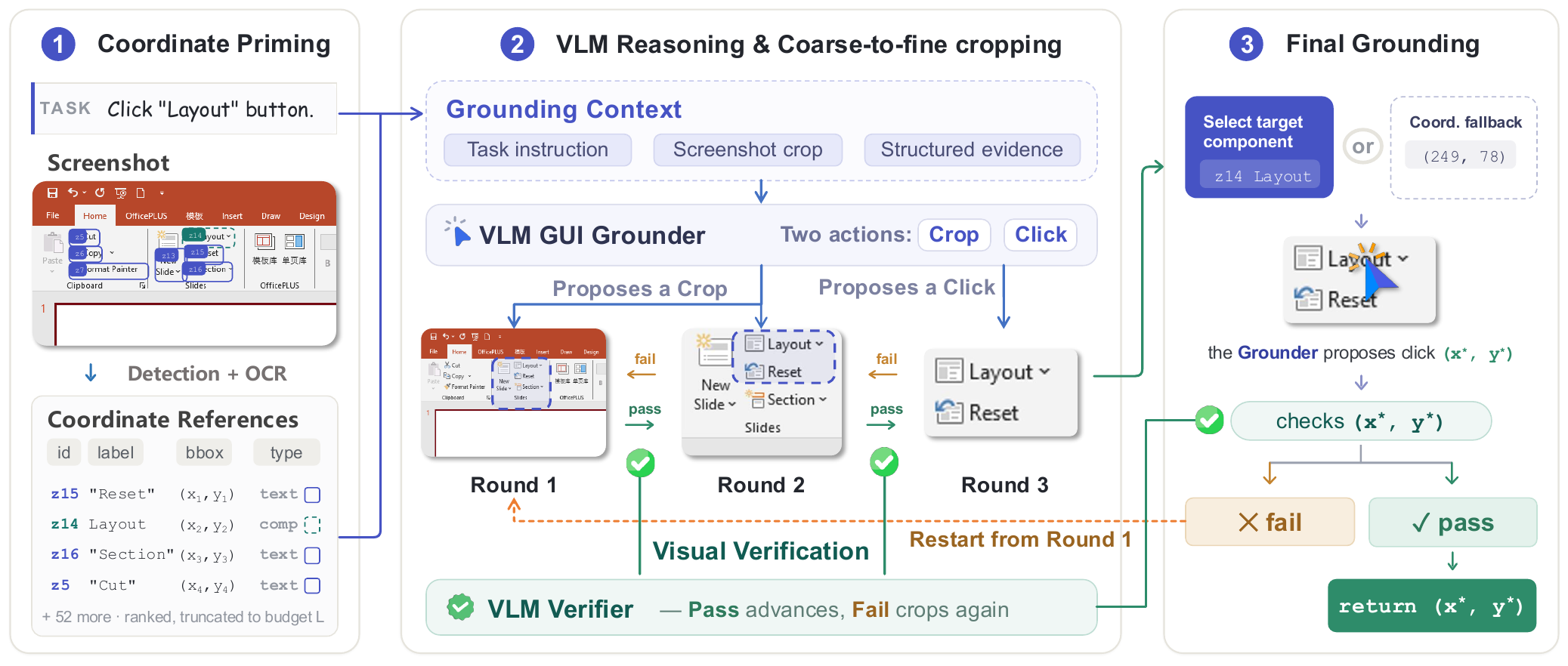}
    \caption{Overview of \ours. Coordinate Priming converts OCR text and detected UI components into coordinate references. Using the task, current crop, and visible references, the VLM either selects a finer crop or proposes a click. Accepted crops advance the process, rejected proposals restart it, and the accepted final click is mapped back to the original screen.}
    \label{fig:gui_grounding_pipeline}
\end{figure*}

Based on this observation, we propose \ours, a coarse-to-fine GUI grounding framework that replaces direct click prediction with a sequence of model-selected visual observations.
As illustrated in Figure~\ref{fig:gui_grounding_pipeline}, \ours primarily includes three components:
1) Coordinate priming converts OCR text and detected UI components into coordinate references containing labels, bounding boxes, and source types.
Presented alongside the screenshot, these references associate visible interface content with explicit screen locations.
2) Coarse-to-fine cropping lets the VLM select the region and scale of the next observation based on the instruction, current crop, and visible references.
Each selected crop provides a more focused view while preserving the surrounding context needed to distinguish small or similar controls.
3) Visual verification checks proposed crops and clicks against the instruction.
Rejected proposals restore the full-screen observation and restart cropping.
Once the final target is accepted, its local position is mapped to the original screen coordinates.
The resulting framework makes the following contributions:
\begin{itemize}
\item We formulate GUI grounding as a coarse-to-fine visual reasoning process in which the VLM selects increasingly focused observations before producing the final coordinate for the requested interface control.
\item We propose \ours, a unified framework that combines coordinate references, instruction-conditioned verification, and model-controlled coarse-to-fine cropping for general-purpose VLM grounding.
\item Experiments across five grounding benchmarks and six general-purpose base models show consistent improvements, by up to $9.3\times$ on ScreenSpot-Pro.
\end{itemize}

\section{Method}
\label{sec:method}

In this section, we first formulate GUI grounding and then describe the three components of \ours: Coordinate Priming, Coarse-to-Fine Cropping, and Visual Verification.
Figure~\ref{fig:gui_grounding_pipeline} illustrates how these components convert a screenshot and instruction into a click coordinate in the original screen.

\subsection{Problem Formulation}
\label{sec:problem_formulation}

Following the standard screen-coordinate formulation~\citep{screenspotpro,screenspotv2}, we model GUI grounding as sequential visual localization over focused views rather than one-shot prediction.
Given a GUI screenshot $I \in \mathbb{R}^{H \times W \times 3}$ with screen domain $\Omega_I=[0,W)\times[0,H)$ and an instruction $g$, GUI grounding predicts the target click point $\hat{p}$ in this domain:
\begin{equation}
    \mathcal{G}(I,g)=\hat{p}\in\Omega_I.
\label{eq:problem_formulation}
\end{equation}
Let $b^{\star}\subseteq\Omega_I$ denote the ground-truth target box, which remains unavailable during inference.
Evaluation counts a prediction as correct when $\hat{p}\in b^{\star}$.

Rather than predicting $\hat{p}$ directly from the full screenshot, \ours constructs successively focused visual observations through cropping.
Each accepted crop becomes the input to the next step, while the final local click is mapped back to the original screen after the target has been determined.

\subsection{Coordinate Priming}
\label{sec:coordinate_priming}

Detected or indexed UI regions can support spatial reasoning~\citep{setofmark,omniparser,gui_actor}, but their locations are not naturally available to a VLM as structured input.
Before each crop selection, Coordinate Priming associates recognized interface content with approximate positions on the current screen during grounding.

Given $I$, an OCR module $\mathcal{O}$ extracts text references such as menu labels, buttons, input hints, and dialog text.
A UI component detector $\mathcal{D}$ extracts component references such as icons, panels, and clickable regions that may not contain readable text.
The two sources are complementary: OCR provides semantic cues for text-described targets, while the detector covers icons and controls without readable labels.
Together, they augment the screenshot with structured spatial information and form the coordinate reference set
\begin{equation}
    R(I)
    = \mathcal{O}(I) \cup \mathcal{D}(I)
    = \{r_k\}_{k=1}^{K}.
\label{eq:coordinate_priming}
\end{equation}
Each reference $r_k=(\mathrm{id}_k,\ell_k,b_k,s_k)$ contains an identifier, a label, an original-screen bounding box, and a source type $s_k\in\{\texttt{text},\texttt{component}\}$.
The set is serialized into the prompt so that visible content can be associated with its approximate screen location during crop selection.

Coordinate references guide localization without defining a closed set of target regions.
The VLM still determines each crop and final click from the current image, while the references help distinguish text labels, similar icons, and nearby clickable areas.
This design preserves direct visual reasoning when detector boxes are imprecise or the intended target differs from the nearest recognized text region.

\subsection{Coarse-to-Fine Cropping}
\label{sec:coarse_to_fine}

Small, densely arranged targets motivate enlarged local views~\citep{screenspotpro,zoomclick,adazoom_gui}, but click-centered zooming can exclude the target or necessary context when its initial point is inaccurate.
Coarse-to-Fine Cropping instead lets the VLM determine the region and scale of each successive observation.

At round $r$, the framework selects references whose original-screen boxes overlap the current crop:
\begin{equation}
    R_r =
    \{r_k \in R(I) \mid b_k \cap B_r \neq \emptyset\}.
\label{eq:visible_evidence}
\end{equation}
The set $R_r$ aligns the prompt with $I_r$ by retaining visible references and removing those outside $B_r$.

Given $(g,I_r,R_r,r)$, the VLM selects an action $a_r$ and produces its corresponding spatial proposal $z_r$:
\begin{equation}
    (a_r,z_r)
    = \mathcal{M}(g,I_r,R_r,r).
\label{eq:crop_decision}
\end{equation}
Here, $a_r\in\{\texttt{crop},\texttt{click}\}$ determines how $z_r$ is interpreted.
For $a_r=\texttt{crop}$, $z_r=\hat{C}_r\subseteq\Omega_r$ is a proposed crop region.
For $a_r=\texttt{click}$, $z_r=\hat{p}_r\in\Omega_r$ is a proposed local click point.
For a click action, selecting a coordinate reference or using the coordinate fallback is resolved into $\hat{p}_r$ before verification.
The proposal is applied only after the Visual Verification stage described below.

Rather than producing the final screen coordinate, a crop action specifies the image region to be examined next.
Early crops preserve broad layout context, while later crops focus on the target component.
Because the VLM selects both region and scale from the current image, the next observation is not fixed around a provisional click.
Within the refinement budget, this process continues until the VLM proposes a local click that passes verification and can be mapped to the original screen coordinate system.

\begin{table*}[t]
\centering
\caption{ScreenSpot-Pro results grouped by base model. Training-based methods train on GUI data, and training-free methods use the base model as is. Imp.\ is the average gain over Direct with the same base model, and all gains of \ours are significant under a two-sided McNemar test ($p<0.05$). Best training-free results within each base model are bolded.}
\label{tab:screenspot_pro_comparison}
\resizebox{\textwidth}{!}{%
\begin{tabular}{@{}llcccccccccccccccc@{}}
\toprule
\multirow{2}{*}[-0.5ex]{\textbf{Base Model}} & \multirow{2}{*}[-0.5ex]{\textbf{Method}} & \multicolumn{2}{c}{\textbf{CAD}} & \multicolumn{2}{c}{\textbf{Dev}} & \multicolumn{2}{c}{\textbf{Creative}} & \multicolumn{2}{c}{\textbf{Scientific}} & \multicolumn{2}{c}{\textbf{Office}} & \multicolumn{2}{c}{\textbf{OS}} & \multicolumn{3}{c}{\textbf{Average}}          & \multirow{2}{*}[-0.5ex]{\textbf{Imp.}} \\ \cmidrule(lr){3-4} \cmidrule(lr){5-6} \cmidrule(lr){7-8} \cmidrule(lr){9-10} \cmidrule(lr){11-12} \cmidrule(lr){13-14} \cmidrule(lr){15-17}
                                     &                                  & Text            & Icon           & Text            & Icon           & Text              & Icon              & Text               & Icon               & Text             & Icon             & Text           & Icon           & Text          & Icon          & \textbf{Avg.} &                                        \\ \midrule[\heavyrulewidth]
\multicolumn{18}{c}{\textbf{Training-based}}                                                                                                                                                                                                                                                                                                                                                     \\ \midrule[\heavyrulewidth]
\multirow{2}{*}{\textbf{Qwen2.5-VL-7B}}       & Direct                           & 16.8 & 1.6 & 46.8 & 4.1 & 35.9 & 7.7 & 49.3 & 7.3 & 52.5 & 20.8 & 37.4 & 6.7 & 38.9 & 7.1 & 26.8 & -- \\
                                     & GUI-Actor-7B                     & 47.7 & 9.4 & 59.1 & 15.9 & 59.6 & 16.1 & 70.1 & 25.5 & 69.5 & 41.5 & 55.1 & 19.1 & 60.0 & 19.7 & 44.6 & +17.8 \\ \midrule
\multirow{2}{*}{\textbf{Qwen3-VL-8B}}         & Direct                           & 46.7 & 10.9 & 79.2 & 23.4 & 68.2 & 14.0 & 73.6 & 30.0 & 76.3 & 30.2 & 65.4 & 21.3 & 67.6 & 21.3 & 49.9 & -- \\
                                     & MAI-UI-8B                        & 73.1 & 42.2 & 85.1 & 51.0 & 76.3 & 29.4 & 77.1 & 36.4 & 89.3 & 56.6 & 80.4 & 49.4 & 79.9 & 42.5 & 65.7 & +15.8 \\ \midrule
\multirow{2}{*}{\textbf{Qwen3-VL-30B-A3B}}    & Direct                           & 51.8 & 15.6 & 76.0 & 24.8 & 69.2 & 20.3 & 76.4 & 27.3 & 80.8 & 37.7 & 75.7 & 38.2 & 70.6 & 26.3 & 53.7 & -- \\
                                     & UI-Venus-1.5-30B-A3B             & 70.6 & 40.6 & 87.7 & 57.9 & 75.8 & 41.3 & 84.0 & 47.3 & 89.8 & 69.8 & 83.2 & 56.2 & 81.2 & 51.0 & 69.6 & +15.9 \\ \midrule
\multirow{2}{*}{\textbf{GPT-4o}}              & Direct                           & 2.0 & 0.0 & 1.3 & 0.0 & 1.0 & 0.0 & 2.1 & 0.0 & 1.1 & 0.0 & 0.0 & 0.0 & 1.3 & 0.0 & 0.8 & -- \\
                                     & OmniParser V2                    & 52.3 & 9.4 & 64.9 & 8.3 & 52.0 & 7.0 & 50.0 & 15.5 & 66.7 & 22.6 & 55.1 & 14.6 & 56.8 & 11.6 & 39.5 & +38.7 \\ \midrule[\heavyrulewidth]
\multicolumn{18}{c}{\textbf{Training-free}}                                                                                                                                                                                                                                                                                                                                                      \\ \midrule[\heavyrulewidth]
\multirow{4}{*}{\textbf{MiniMax-M3.1}} & Direct & 31.0 & 20.3 & 62.3 & 23.4 & 59.1 & 20.3 & 79.2 & 30.9 & 73.4 & 52.8 & 43.9 & 25.8 & 57.8 & 26.7 & 45.9 & -- \\
                                     & AdaZoom-GUI & 58.9 & 39.1 & 81.2 & 49.7 & 74.2 & 36.4 & 67.4 & 35.5 & 84.2 & 73.6 & 77.6 & 46.1 & 73.4 & 44.4 & 62.3 & +16.4 \\
                                     & ZoomClick & 59.4 & 43.8 & 87.7 & 62.1 & 77.3 & \textbf{54.5} & 82.6 & 50.0 & 81.9 & 69.8 & 81.3 & 53.9 & 77.4 & 55.6 & 69.1 & +23.1 \\
                                     & \textbf{\ours} & \textbf{82.2} & \textbf{53.1} & \textbf{92.9} & \textbf{72.4} & \textbf{85.4} & 51.7 & \textbf{88.9} & \textbf{59.1} & \textbf{95.5} & \textbf{83.0} & \textbf{84.1} & \textbf{71.9} & \textbf{88.1} & \textbf{63.9} & \textbf{78.9} & \textbf{+33.0} \\ \midrule
\multirow{6}{*}{\textbf{DeepSeek-V4.1-Flash}} & Direct & 15.2 & 3.1 & 40.9 & 5.5 & 39.9 & 8.4 & 56.9 & 17.3 & 44.1 & 17.0 & 29.9 & 6.7 & 37.3 & 9.3 & 26.6 & -- \\
                                     & BAMI & 45.7 & 20.3 & 73.4 & 29.7 & 69.7 & 30.1 & 82.6 & 37.3 & 79.7 & 50.9 & 63.6 & 37.1 & 68.5 & 33.1 & 55.0 & +28.4 \\
                                     & AdaZoom-GUI & 20.3 & 12.5 & 33.8 & 14.5 & 46.5 & 13.3 & 44.4 & 20.0 & 38.4 & 15.1 & 28.0 & 12.4 & 35.4 & 14.7 & 27.5 & +0.9 \\
                                     & ZoomClick & 58.9 & 21.9 & 77.3 & 32.4 & 70.2 & 37.1 & 75.7 & 37.3 & 85.9 & 49.1 & 72.0 & 37.1 & 72.9 & 35.4 & 58.6 & +32.0 \\
                                     & UI-Zoomer & 50.8 & 23.4 & 74.7 & 33.8 & 68.2 & 38.5 & 68.1 & 28.2 & 75.1 & 50.9 & 70.1 & 51.7 & 67.1 & 36.9 & 55.6 & +29.0 \\
                                     & \textbf{\ours}                   & \textbf{84.3} & \textbf{48.4} & \textbf{91.6} & \textbf{46.2} & \textbf{87.9} & \textbf{43.4} & \textbf{88.2} & \textbf{48.2} & \textbf{91.5} & \textbf{62.3} & \textbf{78.5} & \textbf{59.6} & \textbf{87.4} & \textbf{49.5} & \textbf{72.9} & \textbf{+46.3} \\ \midrule
\multirow{6}{*}{\textbf{GLM-5.3-Flash}}       & Direct & 18.3 & 7.8 & 13.6 & 6.9 & 21.7 & 8.4 & 42.4 & 13.6 & 35.6 & 17.0 & 10.3 & 3.4 & 24.1 & 8.9 & 18.3 & -- \\
                                     & BAMI & 38.6 & 14.1 & 66.2 & 31.0 & 69.2 & 38.5 & 52.8 & 28.2 & 72.9 & 43.4 & 62.6 & 43.8 & 60.1 & 33.4 & 49.9 & +31.6 \\
                                     & AdaZoom-GUI & 53.8 & 35.9 & 60.4 & 26.9 & 61.1 & 37.8 & 84.0 & 45.5 & 74.0 & 54.7 & 39.3 & 16.9 & 62.8 & 34.8 & 52.1 & +33.8 \\
                                     & ZoomClick & 26.4 & 4.7 & 55.2 & 20.0 & 50.0 & 25.9 & 36.1 & 20.0 & 56.5 & 28.3 & 65.4 & 41.6 & 46.9 & 23.7 & 38.0 & +19.7 \\
                                     & UI-Zoomer & 56.3 & 25.0 & 68.2 & 33.8 & 62.1 & 39.2 & 56.2 & 35.5 & 71.8 & 56.6 & 67.3 & 49.4 & 63.4 & 38.7 & 54.0 & +35.7 \\
                                     & \textbf{\ours} & \textbf{85.8} & \textbf{57.8} & \textbf{89.0} & \textbf{52.4} & \textbf{86.4} & \textbf{51.7} & \textbf{91.0} & \textbf{56.4} & \textbf{97.2} & \textbf{84.9} & \textbf{86.0} & \textbf{67.4} & \textbf{89.3} & \textbf{58.6} & \textbf{77.5} & \textbf{+59.2} \\ \midrule
\multirow{2}{*}{\textbf{Gemma-4-31B}}           & Direct & 3.6 & 3.1 & 7.8 & 0.7 & 12.1 & 1.4 & 25.0 & 3.6 & 9.0 & 7.5 & 8.4 & 1.1 & 10.6 & 2.3 & 7.5 & -- \\
                                     & \textbf{\ours} & \textbf{77.7} & \textbf{29.7} & \textbf{86.4} & \textbf{41.4} & \textbf{83.3} & \textbf{38.5} & \textbf{93.1} & \textbf{46.4} & \textbf{92.1} & \textbf{60.4} & \textbf{81.3} & \textbf{52.8} & \textbf{85.5} & \textbf{43.7} & \textbf{69.5} & \textbf{+62.0} \\ \midrule
\multirow{2}{*}{\textbf{Claude Opus 4.7}}     & Direct                           & 36.5 & 20.3 & 84.4 & 45.5 & 77.3 & 35.7 & 81.9 & 43.6 & 75.1 & 47.2 & 61.7 & 36.0 & 68.8 & 38.9 & 57.4 & -- \\
                                     & \textbf{\ours}                   & \textbf{85.3} & \textbf{68.8} & \textbf{94.2} & \textbf{69.0} & \textbf{89.9} & \textbf{69.2} & \textbf{91.0} & \textbf{61.8} & \textbf{95.5} & \textbf{88.7} & \textbf{87.9} & \textbf{65.2} & \textbf{90.6} & \textbf{68.9} & \textbf{82.3} & \textbf{+24.9} \\ \midrule
\multirow{2}{*}{\textbf{GPT-5.5}}             & Direct                           & 84.3 & 70.3 & 72.7 & 53.1 & 80.8 & 66.4 & 91.7 & 60.9 & 89.8 & 81.1 & 65.4 & 62.9 & 81.8 & 63.4 & 74.8 & -- \\
                                     & \textbf{\ours}                   & \textbf{90.9} & \textbf{71.9} & \textbf{95.5} & \textbf{78.6} & \textbf{92.9} & \textbf{74.8} & \textbf{93.8} & \textbf{72.7} & \textbf{98.3} & \textbf{88.7} & \textbf{94.4} & \textbf{85.4} & \textbf{94.2} & \textbf{77.8} & \textbf{87.9} & \textbf{+13.1} \\ \bottomrule
\end{tabular}%
}
\end{table*}

\subsection{Visual Verification}
\label{sec:visual_verification}

Even a visually plausible crop or click may select a neighboring control in dense interfaces~\citep{mmbench_gui}. GUI-Lens therefore supports instruction-conditioned verification as an optional refinement to the grounding prediction. 

Given action $a_r$ and proposal $z_r$, the framework marks the proposed box or point on the current observation and submits the marked image to a verification prompt:
\begin{equation}
\begin{aligned}
    \tilde{I}_r
    &= \operatorname{Mark}(I_r,z_r,a_r), \\
    v_r
    &= \mathcal{V}(g,\tilde{I}_r,a_r).
\end{aligned}
\label{eq:visual_verification}
\end{equation}
Here, $v_r\in\{\texttt{accept},\texttt{reject}\}$ is the verification decision.
The grounding call $\mathcal{M}$ and verification call $\mathcal{V}$ use the same VLM backend with stage-specific prompts.
For a crop proposal, verification checks whether the marked region contains the requested target.
For a click proposal, it checks whether the marked point identifies the requested control.

An accepted crop updates the next screen region as
\begin{equation}
    B_{r+1}
    = \operatorname{Pad}\bigl(\Phi_r(\hat{C}_r)\bigr)
      \cap \Omega_I.
\label{eq:crop_update}
\end{equation}
Here, the local crop is mapped to the original screen, padded to retain surrounding context, and clipped to the screen boundary.
The resulting region preserves the selected context while exposing the target at a finer scale for the next view.

When a click proposal is accepted, its local point is mapped directly to the original screen coordinate system:
\begin{equation}
    \hat{p}=\Phi_r(\hat{p}_r).
\label{eq:final_click}
\end{equation}
Rejected proposals restore $B_0$ and restart cropping from the full-screen observation.
The framework returns $\hat{p}$ only after the final marked click is accepted.
Intermediate crop regions construct subsequent image observations, whereas the framework returns $\hat{p}$ as a single click coordinate in the original screen coordinate system used by the environment.


\begin{table*}[t]
\centering
\caption{Performance comparison on UI-Vision and MMBench-GUI-L2. Baseline results follow the original benchmark reports~\citep{ui_vision,mmbench_gui}. Best and second-best scores are bolded and underlined, respectively. Improvements of \ours over the corresponding single-shot baselines are statistically significant under a two-sided McNemar test ($p<0.05$).}
\label{tab:mmbench_ui_vision}
\resizebox{\textwidth}{!}{%
\begin{tabular}{l*{17}{c}}
\toprule
\textbf{Model / Method}
& \multicolumn{4}{c}{\textbf{UI-Vision}}
& \multicolumn{13}{c}{\textbf{MMBench-GUI-L2}} \\
\cmidrule(lr){2-5}\cmidrule(lr){6-18}
& & & &
& \multicolumn{2}{c}{\textbf{Win.}}
& \multicolumn{2}{c}{\textbf{macOS}}
& \multicolumn{2}{c}{\textbf{Linux}}
& \multicolumn{2}{c}{\textbf{iOS}}
& \multicolumn{2}{c}{\textbf{Android}}
& \multicolumn{2}{c}{\textbf{Web}}
& \\
\cmidrule(lr){6-7}\cmidrule(lr){8-9}\cmidrule(lr){10-11}
\cmidrule(lr){12-13}\cmidrule(lr){14-15}\cmidrule(lr){16-17}
& Basic & Func. & Spatial & Avg.
& Basic & Adv.
& Basic & Adv.
& Basic & Adv.
& Basic & Adv.
& Basic & Adv.
& Basic & Adv.
& Avg. \\
\midrule
\multicolumn{18}{c}{General Models} \\
\midrule
GPT-4o
& 1.58 & 1.52 & 1.03 & 1.38
& 1.48 & 1.10 & 8.69 & 4.34 & 1.05 & 1.02
& 5.10 & 3.33 & 2.53 & 1.41 & 3.23 & 2.92 & 2.87 \\
Claude-3.7-Sonnet
& 9.48 & 7.73 & 7.60 & 8.27
& 1.48 & 0.74 & 12.46 & 7.51 & 1.05 & 0.00
& 13.69 & 10.61 & 1.40 & 1.40 & 3.23 & 2.27 & 4.66 \\
Qwen2.5-VL-7B
& 1.24 & 0.79 & 0.51 & 0.85
& 31.37 & 16.54 & 31.30 & 21.97 & 21.47 & 12.24
& 66.56 & 55.15 & 35.11 & 35.21 & 40.32 & 32.47 & 33.85 \\
\midrule
\multicolumn{18}{c}{Specialized GUI Models} \\
\midrule
ShowUI-2B
& 8.07 & 7.67 & 2.07 & 5.94
& 9.23 & 4.41 & 24.06 & 10.40 & 25.13 & 11.73
& 28.98 & 19.70 & 17.42 & 8.73 & 22.90 & 12.66 & 15.96 \\
OS-Atlas-Base-7B
& 12.20 & 11.20 & 3.67 & 9.02
& 36.90 & 18.75 & 44.35 & 21.68 & 31.41 & 13.27
& 74.84 & 48.79 & 69.60 & 46.76 & 61.29 & 35.39 & 41.42 \\
AGUVIS-7B
& 17.80 & 18.30 & 5.06 & 13.70
& 37.27 & 21.69 & 48.12 & 33.27 & 33.51 & 25.00
& 67.52 & 65.15 & 60.96 & 50.99 & 61.61 & 45.45 & 45.66 \\
UGround-V1-7B
& 15.40 & 17.10 & 6.25 & 12.90
& 66.79 & 38.97 & 71.30 & 48.55 & 56.54 & 31.12
& \underline{92.68} & 70.91 & \underline{93.54} & 70.99
& \underline{88.71} & 64.61 & 65.68 \\
UI-TARS-72B-DPO
& \underline{31.40} & \underline{30.50} & \underline{14.70} & \underline{25.50}
& \underline{78.60} & \underline{51.84}
& \underline{80.29} & \underline{62.72}
& \underline{68.59} & \underline{51.53}
& 90.76 & \underline{81.21}
& 92.98 & \underline{80.00}
& 88.06 & \underline{68.51}
& \underline{74.25} \\
\midrule
\textbf{\ours (GPT-5.5)}
& \textbf{73.08} & \textbf{67.04} & \textbf{66.05} & \textbf{68.64}
& \textbf{94.03} & \textbf{83.46}
& \textbf{91.86} & \textbf{86.73}
& \textbf{89.01} & \textbf{78.57}
& \textbf{97.45} & \textbf{92.73}
& \textbf{97.47} & \textbf{94.37}
& \textbf{97.09} & \textbf{87.91}
& \textbf{91.52} \\
\bottomrule
\end{tabular}%
}
\end{table*}

\section{Experiments}
\label{sec:experiments}

\subsection{Experimental Settings}
\label{sec:exp_settings}

\paragraph{Datasets.} Our evaluation employs two categories of datasets: static GUI grounding benchmarks and interactive computer-use tasks.
Static grounding is evaluated on ScreenSpot-Pro~\citep{screenspotpro}, ScreenSpot-v2~\citep{screenspotv2}, MMBench-GUI-L2~\citep{mmbench_gui}, UI-Vision~\citep{ui_vision}, and OSWorld-G~\citep{osworld_g}, which collectively cover professional applications, multiple platforms, and diverse target types.
Interactive performance is evaluated on OSWorld~\citep{osworld}, where agents complete tasks in real applications.

\paragraph{Evaluation Metrics.} For ScreenSpot-Pro, ScreenSpot-v2, MMBench-GUI-L2, and UI-Vision Element Grounding, we report point-in-box accuracy, counting a prediction as correct if the predicted click point falls within the ground-truth bounding box.
OSWorld-G additionally counts a polygon target as correct when the point falls inside the polygon; its 54 refusal examples, which have no target on screen, are left out.
We provide benchmark-specific breakdowns and report OSWorld task scores overall and by domain.

\paragraph{Baselines.} Comparisons include representative general VLMs, specialized GUI models, and GUI grounding systems.
The corresponding result tables provide the comparison details.
Their methodological distinctions are discussed in Section~\ref{sec:related}.
For MiniMax-M3.1, DeepSeek-V4.1-Flash, and GLM-5.3-Flash, the single-shot baselines, BAMI, AdaZoom-GUI, ZoomClick, and UI-Zoomer~\citep{ui_zoomer} are evaluated by us; other baseline results follow official leaderboards or the corresponding papers.
Descriptions of the evaluated baseline systems and their comparison settings are provided in Appendix~\ref{app:baselines}.

\paragraph{Implementation Details.} All experiments are implemented in GUI-Agent-Harness.
ScreenSpot-Pro is evaluated with six base models, namely GPT-5.5, Claude Opus 4.7, MiniMax-M3.1, DeepSeek-V4.1-Flash, GLM-5.3-Flash, and Gemma-4-31B~\citep{gpt55,deepseek_v41_flash,glm53_flash,gemma4}; OSWorld-G with MiniMax-M3.1, DeepSeek-V4.1-Flash, and GLM-5.3-Flash; and ScreenSpot-v2, MMBench-GUI-L2, and UI-Vision with GPT-5.5.
Each base model is called through one endpoint with its default image budget, which is 10.24, 1.81, 4.07, and 6.21 megapixels for GPT-5.5, DeepSeek-V4.1-Flash, MiniMax-M3.1, and GLM-5.3-Flash, about 0.65 megapixels for Gemma-4-31B, and a long edge of 2{,}000 pixels for Claude Opus 4.7.
For OSWorld, Claude Opus 4.7 acts in a $1920\times1080$ Ubuntu virtual machine with 15 interaction steps allowed per task.

GPA-GUI-Detector~\citep{gpa-gui-detector} detects UI components with a confidence threshold of 0.1 and an NMS IoU threshold of 0.3, and EasyOCR~\citep{baek2019craft,shi2017crnn} recognizes English and Simplified Chinese text.
Cropping runs for at most eight rounds, or five for MiniMax-M3.1, and intermediate and final crops are enlarged by at most $5\times$ and $8\times$.
Verification allows up to six retries, or three for MiniMax-M3.1.
Crop proposal, verification, and final grounding use separate calls to the same base model.
Appendices~\ref{app:grounding_ablation} and~\ref{app:implementation} give the full configuration.

\begin{table}[t]
\centering
\caption{OSWorld-G results grouped by base model.}
\label{tab:osworld_g}
\resizebox{\columnwidth}{!}{%
\begin{tabular}{@{}llccccc@{}}
\toprule
\textbf{Base Model} & \textbf{Method} & \textbf{Text} & \textbf{Elem.} & \textbf{Layout} & \textbf{Manip.} & \textbf{Avg.} \\
\midrule
\multirow{6}{*}{\textbf{MiniMax-M3.1}} & Direct & 85.8 & 84.0 & \textbf{86.2} & 77.3 & 81.8 \\
 & BAMI & 78.3 & 82.4 & 82.4 & 65.9 & 76.9 \\
 & AdaZoom-GUI & 86.7 & 84.0 & 82.4 & 78.8 & 83.1 \\
 & ZoomClick & 78.8 & 81.0 & 79.5 & 66.7 & 76.7 \\
 & UI-Zoomer & \textbf{88.8} & \textbf{86.3} & 84.5 & 80.3 & \textbf{84.3} \\
 & \textbf{\ours} & 87.1 & 84.0 & 84.5 & \textbf{81.1} & 83.3 \\
\midrule
\multirow{6}{*}{\textbf{DeepSeek-V4.1-Flash}} & Direct & 69.6 & 53.6 & 57.7 & 47.7 & 54.7 \\
 & BAMI & 79.6 & 72.2 & 73.6 & 64.4 & 70.8 \\
 & AdaZoom-GUI & 45.8 & 37.6 & 45.2 & 29.5 & 37.3 \\
 & ZoomClick & 56.2 & 50.3 & 52.7 & 43.2 & 50.6 \\
 & UI-Zoomer & 69.6 & 69.3 & 69.9 & 51.5 & 63.9 \\
 & \textbf{\ours} & \textbf{86.7} & \textbf{84.3} & \textbf{85.4} & \textbf{80.3} & \textbf{82.7} \\
\midrule
\multirow{6}{*}{\textbf{GLM-5.3-Flash}} & Direct & 74.2 & 70.9 & 75.7 & 54.5 & 68.0 \\
 & BAMI & 62.5 & 61.8 & 59.4 & 46.2 & 57.3 \\
 & AdaZoom-GUI & 85.8 & 81.0 & 83.7 & 73.5 & 79.6 \\
 & ZoomClick & 42.9 & 49.3 & 46.4 & 34.8 & 43.5 \\
 & UI-Zoomer & 61.7 & 68.3 & 66.5 & 40.2 & 61.0 \\
 & \textbf{\ours} & \textbf{88.3} & \textbf{88.2} & \textbf{87.9} & \textbf{78.8} & \textbf{85.3} \\
\bottomrule
\end{tabular}%
}
\end{table}

\subsection{Main Results: Grounding Performance}
\label{sec:exp_main}

In this section, we evaluate \ours on ScreenSpot-Pro, UI-Vision, and MMBench-GUI-L2 to examine whether its improvements persist across base models, target types, platforms, and difficulty levels.
Tables~\ref{tab:screenspot_pro_comparison} and \ref{tab:mmbench_ui_vision} report same-base-model and cross-benchmark comparisons across the evaluated grounding settings.
Appendix~\ref{app:per_domain} reports detailed ScreenSpot-v2 results across platforms and target types.

\textbf{Consistent gains across base models.}
\ours improves all six base models on ScreenSpot-Pro, and the gain grows as direct clicking gets weaker.
Gemma-4-31B sees each screenshot at about 0.65 megapixels and clicks correctly on 7.5\% of examples, while \ours raises it to 69.5\%; GLM-5.3-Flash rises from 18.3\% to 77.5\% and DeepSeek-V4.1-Flash from 26.6\% to 72.9\%.
These base models recognize the requested control but see it at too low a resolution to place the click, and each crop enlarges the target until its position becomes clear.
Text targets gain from the recognized labels in the coordinate references, while icons carry no text cue and depend on the enlarged view and the visual recognition of the base model, so Icon accuracy rises on every base model yet stays below Text accuracy.

\textbf{Advantage over zoom baselines.}
\ours exceeds the strongest training-free baseline of each base model by 9.8 points on MiniMax-M3.1 and 14.3 points on DeepSeek-V4.1-Flash, both over ZoomClick, and by 23.5 points on GLM-5.3-Flash over UI-Zoomer.
The baselines also depend on the base model: AdaZoom-GUI adds 33.8 points on GLM-5.3-Flash and only 0.9 points on DeepSeek-V4.1-Flash.
Most of these baselines refine around an early prediction on the full screenshot, so an inaccurate first estimate carries into the final click.
The progressively narrowed crops of \ours only need to contain the target, and verification returns to a broader view when a crop misses it, which keeps the gain stable across all base models and screen layouts.

\textbf{Robustness to harder instructions.}
On UI-Vision and MMBench-GUI-L2, \ours with GPT-5.5 gives the best result in every column, reaching 68.64\% and 91.52\% on average against 25.50\% and 74.25\% for UI-TARS-72B-DPO.
Its accuracy drops by 7.0 points from Basic to Spatial on UI-Vision, while UI-TARS-72B-DPO loses more than half of its Basic accuracy, and the average gap between Basic and Advanced on MMBench-GUI-L2 shrinks from 17.2 points to 7.2 points.
Spatial and Advanced instructions identify the target through its neighbors or its function, which calls for reasoning over the layout.
\ours leaves this reasoning to a general-purpose VLM and gives it enlarged views of candidate regions with the coordinates of nearby text and components, so its reasoning rests on visual evidence.

\begin{table}[t]
\centering
\caption{OSWorld results, with baselines from the official OSWorld-Verified leaderboard~\citep{osworld_leaderboard}.}
\label{tab:baseline_comparison}
\resizebox{\linewidth}{!}{%
\begin{tabular}{lcccc}
\toprule
\textbf{Method (Model)}
& \textbf{Chrome}
& \textbf{Multi-App}
& \textbf{OS}
& \textbf{Avg.} \\
\midrule
Claude Sonnet 4.6 & 78.5 & 60.2 & 91.7 & 70.1 \\
Claude Opus 5 & \underline{80.2} & 76.8 & 79.2 & 78.1 \\
Claude Fable 5 & 78.1 & \underline{79.3} & \underline{95.8} & \underline{81.4} \\
Kimi K2.6 & 76.7 & 55.0 & 79.2 & 64.7 \\
MiniMax-M3 & 76.0 & 61.6 & 83.3 & 68.9 \\
Holo3-35B-A3B & 78.3 & 62.9 & \underline{95.8} & 72.1 \\
Muse Spark 1.1 & 73.8 & 71.7 & \underline{95.8} & 75.9 \\
Coasty CUA v1 & 67.3 & 69.2 & \textbf{100.0} & 73.2 \\
GBOX Agent & 63.0 & 49.7 & 70.8 & 56.6 \\
HIPPO Agent (Claude Opus 4.5) & 60.4 & 64.3 & 87.5 & 66.6 \\
VLAA-GUI (Claude Opus 4.5) & 66.6 & 61.1 & 91.7 & 67.2 \\
OpenAPA (Gemini 3.1 Pro) & 71.7 & 65.7 & 83.3 & 70.0 \\
Pointer Agent (Claude Opus 4.7) & 78.2 & 74.4 & \underline{95.8} & 78.6 \\
\midrule
\textbf{\ours (Claude Opus 4.7)} & \textbf{93.5} & \textbf{80.0} & \textbf{100.0} & \textbf{86.8} \\
\bottomrule
\end{tabular}%
}
\end{table}

\begin{figure*}[t]
    \centering
    \begin{minipage}[t]{0.334\textwidth}
        \centering
        \includegraphics[width=\linewidth]{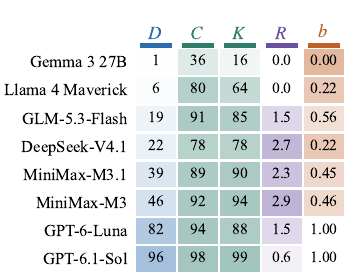}
        \par\textbf{(a)} Probe inputs
    \end{minipage}\hfill
    \begin{minipage}[t]{0.362\textwidth}
        \centering
        \includegraphics[width=\linewidth]{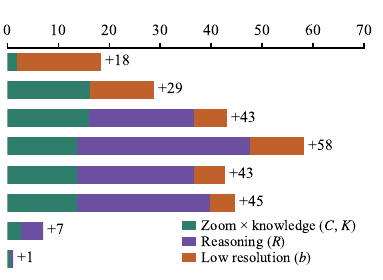}
        \par\textbf{(b)} Expected gain by term (points)
    \end{minipage}\hfill
    \begin{minipage}[t]{0.298\textwidth}
        \centering
        \includegraphics[width=\linewidth]{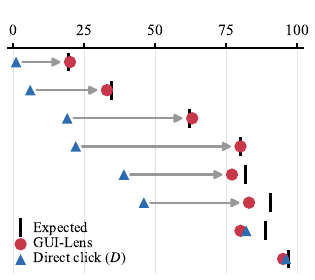}
        \par\textbf{(c)} Accuracy (\%)
    \end{minipage}
    \caption{Probes and gain of coarse-to-fine cropping on eight base models, one row each. (a) Inputs of Eq.~\eqref{eq:gain}, colored by the term they enter. (b) Expected gain by term, base model held out of the fit. (c) Direct click, \ours, and expected accuracy.}
    \label{fig:backbone_analysis}
\end{figure*}

\begin{figure}[t]
    \centering
    \includegraphics[width=\linewidth]{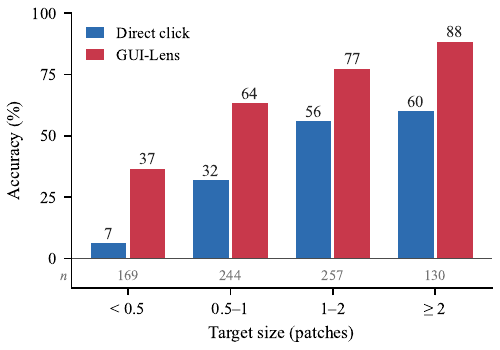}
    \caption{Accuracy of direct clicking and \ours by target size in vision patches, over the eight base models of Figure~\ref{fig:backbone_analysis}.}
    \label{fig:size_gain}
\end{figure}

\begin{figure*}[t]
    \centering
    \begin{minipage}[t]{0.49\textwidth}
        \centering
        \includegraphics[width=\linewidth]{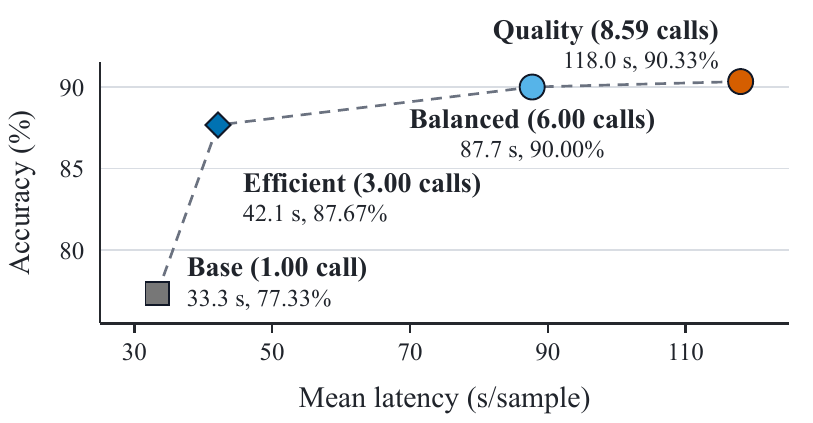}
        \par\textbf{(a)} Accuracy--efficiency trade-off
    \end{minipage}\hfill
    \begin{minipage}[t]{0.49\textwidth}
        \centering
        \includegraphics[width=\linewidth]{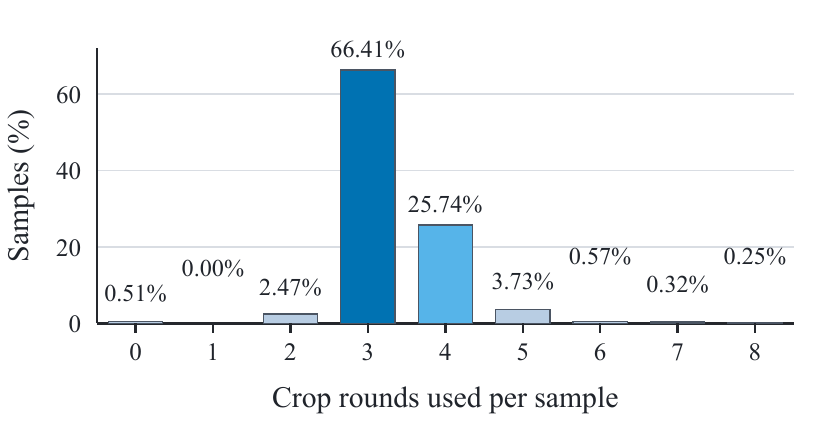}
        \par\textbf{(b)} Crop-round distribution
    \end{minipage}
    \caption{Accuracy--efficiency analysis on ScreenSpot-Pro with GPT-5.5. (a) Accuracy and latency under four inference configurations, with observable VLM invocations shown in parentheses. (b) Distribution of crop rounds used per sample.}
    \label{fig:efficiency}
\end{figure*}

\begin{figure*}[t]
    \centering
    \includegraphics[width=\textwidth]{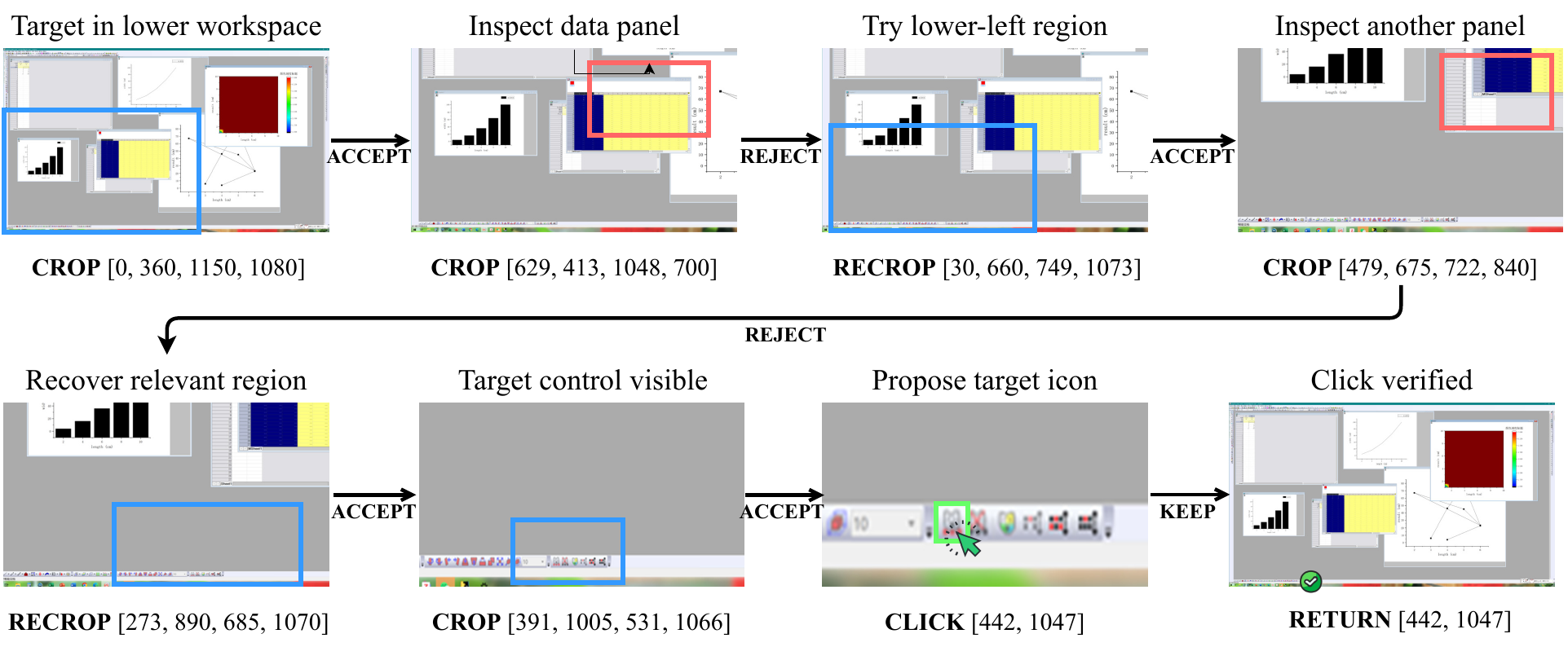}
    \caption{A grounding trajectory of \ours. Verification rejects crop proposals that omit the target, after which the framework selects a new region, progressively exposes the target control, and verifies the final click.}
    \label{fig:case-study}
\end{figure*}

\subsection{Which Base Models Benefit from Coarse-to-Fine Cropping}
\label{sec:exp_backbone}

The gain of coarse-to-fine cropping varies widely across base models, and running the full multi-round procedure on every candidate base model is costly.
We therefore ask whether a few single-call probes can predict, before \ours is run, how much cropping improves over the base model's own direct click.
We study eight base models~\citep{deepseek_v41_flash,glm53_flash,gemma3} whose direct-click accuracy ranges from 1\% to 96\% (Table~\ref{tab:backbone_probes}); each is one model at one reasoning effort through one endpoint, shared by its probes and its \ours run.

\paragraph{Probes and gain model.}
Each probe is one call per example on a stratified 100-example subset of ScreenSpot-Pro (Appendix~\ref{app:backbone}).
$D$: the direct click on the full screenshot hits the target, taking the best of four coordinate formats so that $D$ reflects localization.
$C$: the quarter-screen window centered on that click contains the target.
$K$: the base model picks the target among eight enlarged candidates.
$R=\log_{10}(1+\text{reasoning tokens per call})$ in the knowledge probe.
$b$: the image resolution the endpoint accepts, on a log scale from 0 for the smallest budget we observe (0.80 megapixels) to 1 for an endpoint that keeps every screenshot at its original resolution.
We model the gain on each example and average over examples:
\begin{equation}
\hat{\Delta} = \alpha\,(C-D)\,K + (1-D)\,\big(\gamma R + \rho\,(1-b)\big) .
\label{eq:gain}
\end{equation}
The first term is the room for zooming: the click lands near the target but misses it, and zooming helps when the base model recognizes the enlarged target.
The second applies to examples the direct click misses, which are recovered more often when the base model reasons and when the endpoint has downscaled the screenshot, since a crop restores lost detail.
Least squares over all 800 model--example pairs gives $\alpha=0.31$, $\gamma=0.16$, and $\rho=0.17$ for all models.

\paragraph{Probes predict the gain.}
Fitted without the base model it predicts, Eq.~\eqref{eq:gain} lies within 3.3 points of the measured gain on average (Figure~\ref{fig:backbone_analysis}c), and it sorts the base models into the same three groups as the measurement (Table~\ref{tab:backbone_probes}).
The gain of \ours thus \textbf{follows from a few measurable traits} of the base model.
At 500 calls per base model, the probes tell in advance whether the full multi-round procedure is worth running: it pays off for base models that \textbf{land near the target but miss it}, and it fades as the direct click approaches 80\%.

\paragraph{Which factors matter.}
Figure~\ref{fig:backbone_analysis}b shows which traits produce the gain.
\textbf{Reasoning matters most}: removing $R$ from Eq.~\eqref{eq:gain} raises the held-out error from 3.3 to 12.5 points, more than any other term.
Each round of \ours judges a crop and chooses the next step, and a base model that reasons checks each crop and discards the wrong ones; Llama~4 Maverick and DeepSeek-V4.1 land near the target equally often, yet DeepSeek-V4.1, which reasons at length, gains twice as much.
Zooming works \textbf{near the target}: a crop enlarges the area around the model's own guess, so it fixes clicks that land close to the target, and the GPT models hit most targets directly, which leaves few close misses.
Finally, \textbf{downscaling} by the endpoint discards detail that a crop restores, which explains most of the gain of Gemma~3 27B, the model that sees the screen at the lowest resolution.

\paragraph{Small components need cropping.}
A vision encoder sees the screenshot as patches, and after downscaling many targets cover less than one patch, so they are merged with their surroundings into a single visual token.
Such targets are \textbf{the hardest for direct clicks}: below half a patch, they succeed on 7\% of targets, against about 60\% for targets of one patch or more (Figure~\ref{fig:size_gain}).
Cropping spreads the target over several patches, and \ours recovers about 30 points at every size.
An error of GPT-6.1-Sol shows this directly: ``Chip planner'' is a $23\times23$-pixel icon that Sol misses in all three draws on the original screenshot and hits in all three once the screenshot is enlarged $1.5\times$.
The model knows what to look for, and cropping gives it the resolution to see it.

\subsection{Computer Use}
\label{sec:exp_compute_use}

In this section, we evaluate \ours on desktop computer use at two levels: grounding single instructions on OSWorld-G screenshots, and completing full tasks in the Chrome, Multi-Apps, and OS domains of OSWorld.
These domains cover browser operations, cross-application workflows, and system-level tasks, representing different types of complex interaction in realistic desktop environments.

Table~\ref{tab:osworld_g} shows that \ours gives the \textbf{best average for DeepSeek-V4.1-Flash and GLM-5.3-Flash} and trails UI-Zoomer by 1.0 point on MiniMax-M3.1.
The gain grows as direct clicking gets weaker: DeepSeek-V4.1-Flash rises from 54.7\% to 82.7\% and GLM-5.3-Flash from 68.0\% to 85.3\%, while MiniMax-M3.1 already clicks accurately on these 1080p screenshots and gains 1.5 points.
The zoom baselines are \textbf{base-model dependent}: AdaZoom-GUI helps GLM-5.3-Flash and hurts DeepSeek-V4.1-Flash, BAMI shows the opposite pattern, and ZoomClick falls below direct clicking on all three base models.
Each of these methods commits to the zoom region that the base model picks first, so a wrong choice carries into the final click, whereas the progressively narrowed crops of \ours only need to contain the target.
The largest gains appear on \textbf{fine-grained manipulation}, whose targets such as slider handles span only a few pixels and become clickable once magnified.

Table~\ref{tab:baseline_comparison} demonstrates the \textbf{strong transferability} of \ours to interactive computer-use tasks, with the best average and the best or tied-best performance in all three domains.
Compared with Pointer Agent, which uses the same Claude Opus 4.7 base model, \ours performs better in every domain, indicating that the gain comes from the agent design built around the base model.
Overall, the results show that \ours can be integrated into a complete interactive agent while maintaining stable performance across different computer-use settings and task domains.

\subsection{Efficiency Analysis}
\label{sec:efficiency}

This section evaluates how crop budget and verification affect accuracy and inference overhead on ScreenSpot-Pro with GPT-5.5.
Figure~\ref{fig:efficiency} compares native single-shot inference with Quality (up to four crops with verification), Balanced (up to two crops with verification), and Efficient (up to two crops without verification) on a shared 300-example subset, and reports the crop-round distribution of a separate 1,581-example full-scale run.
Latency covers the full pipeline, while observable VLM invocations measure model overhead because token usage was unavailable for comparison.

Overall, \ours provides a \textbf{competitive accuracy--efficiency trade-off}: Efficient improves single-shot accuracy by 10.34 points at $1.27\times$ the latency, while Balanced remains within 0.33 points of Quality with 25.6\% lower latency and 30.2\% fewer VLM invocations.
In the full-scale run, 95.1\% of samples terminate within four crop rounds and only 0.25\% exhaust the eight-round budget, indicating that most samples do not approach the maximum crop budget.
At equal refinement rounds, coarse-to-fine cropping also outperforms click-centered zooming (Figure~\ref{fig:crop_vs_zoom_rounds}), so the gains do not stem from additional calls alone but from better views.

\subsection{Case Study}
\label{sec:case}

This section illustrates how \ours recovers from inaccurate intermediate crops during grounding.
As shown in Figure~\ref{fig:case-study}, the framework first narrows the search to the lower workspace, but rejects a proposed crop of the data panel because it does not contain the target.
It then examines an alternative panel, rejects that proposal, and restores a broader relevant region rather than committing the localization error.
From this recovered view, successive accepted crops expose the target control at finer resolution, after which the proposed icon is verified and mapped back to the original screen as the final click.
The trajectory demonstrates that an incorrect intermediate crop need not determine the final result because verification can redirect the subsequent visual observations.


\begin{table}[t]
\centering
\caption{Ablation results for \ours on a randomly sampled 300-example subset of ScreenSpot-Pro. $\Delta$ is the change from the full configuration in points, and $^*$ marks a significant change under a two-sided McNemar test ($p<0.05$).}
\label{tab:component_ablation}
\resizebox{\linewidth}{!}{%
\begin{tabular}{@{}lcccccc@{}}
\toprule
& \multicolumn{2}{c}{\textbf{GPT-5.5}}
& \multicolumn{2}{c}{\textbf{\begin{tabular}[b]{@{}c@{}}DeepSeek-\\V4.1-Flash\end{tabular}}}
& \multicolumn{2}{c}{\textbf{MiniMax-M3}} \\
\cmidrule(lr){2-3}\cmidrule(lr){4-5}\cmidrule(l){6-7}
\textbf{Configuration}
& \textbf{Acc.} & $\boldsymbol{\Delta}$
& \textbf{Acc.} & $\boldsymbol{\Delta}$
& \textbf{Acc.} & $\boldsymbol{\Delta}$ \\
\midrule
\textbf{Full \ours}
& \textbf{88.7} & -- & \textbf{71.7} & -- & \textbf{47.7} & -- \\
w/o Coordinate Priming
& 87.7 & $-1.0$ & 57.7 & $-14.0^*$ & 40.3 & $-7.4^*$ \\
w/o Cropping
& 78.3 & $-10.4^*$ & 10.0 & $-61.7^*$ & 32.0 & $-15.7^*$ \\
w/o Visual Verification
& 87.0 & $-1.7$ & 65.7 & $-6.0^*$ & 42.7 & $-5.0^*$ \\
\bottomrule
\end{tabular}
}
\end{table}

\subsection{Ablation Study}
\label{sec:exp_ablation}

This section conducts leave-one-component-out experiments on a randomly sampled 300-example subset of ScreenSpot-Pro to examine the contribution of the three components across different base models.
All variants use the same examples and evaluation setting, as reported in Table~\ref{tab:component_ablation}.

Three findings emerge from the results.
\textbf{Cropping drives the main gains.}
Removing coarse-to-fine cropping causes the largest performance degradation for all three base models, establishing successive selection of focused screen regions as the primary source of improvement.
\textbf{Coordinate priming benefits weaker grounders.}
Removing it causes significant drops for DeepSeek-V4.1-Flash and MiniMax-M3, whereas the 1.0-point change for GPT-5.5 is not significant, so strong base models do not rely on detector output.
\textbf{Verification improves reliability.}
It significantly helps both weaker base models and slightly helps GPT-5.5, whose crop proposals are already accurate before any verification.

Together, the ablations show that coordinate priming and verification support visual refinement, while cropping provides the main source of improvement.

\section{Related Work}
\label{sec:related}
\subsection{Direct Prediction and Spatial Cues}

GUI grounding predicts a click point, box, or action coordinate from a screenshot and instruction.
This formulation is widely used to evaluate general-purpose VLMs~\citep{gpt4o,claude37_sonnet,gpt55,qwen2_vl,qwen25_vl,qwen3_vl}.
GUI-specific models learn localization from grounding or interaction data, including SeeClick, UGround, OS-Atlas, AGUVIS, UI-TARS, ShowUI, GTA1, UI-Venus, MAI-UI, and Holo~\citep{seeclick,uground,screenspotv2,aguvis,uitars,showui,gta1,ui_venus,ui_venus_15,mai_ui,holo2,holo3}.
SeeClick and UGround emphasize cross-interface grounding supervision, whereas OS-Atlas, AGUVIS, and UI-TARS integrate localization with action prediction.
\ours instead refines general-purpose VLM grounding.

Component-selection methods replace open-ended coordinates with structured interface elements.
Set-of-Mark indexes detected regions, OmniParser parses semantic UI elements, and GUI-Actor selects visual patches~\citep{setofmark,omniparser,gui_actor}.
Although these methods provide explicit spatial references, their action spaces depend on candidate coverage, boundaries, and parser quality.
Missing or inaccurate candidates can therefore exclude the intended target before the final selection.
For crop and click decisions, \ours combines the current image with non-exclusive component references.

Attention-based methods derive locations from internal VLM representations.
TAG converts instruction-to-image attention into target locations, while GUI-AIMA aligns intrinsic attention with GUI targets using patch supervision and contextual anchors~\citep{tag,gui_aima}.
These methods show that pretrained VLM representations contain useful spatial information.
Such methods require model-specific representations and an internal-response-to-screen mapping.
\ours instead uses visible, instruction-conditioned evidence through standard VLM interfaces.

\subsection{Inference-Time Visual Refinement}

Inference-time refinement exposes small or dense elements through local views.
DRS-GUI and GUI-ARP select regions, ZoomClick studies multi-step zooming, AdaZoom-GUI refines instructions during zooming, and UI-Zoomer uses prediction uncertainty~\citep{drs_gui,gui_arp,zoomclick,adazoom_gui,ui_zoomer}.
MVP aggregates multiple views, while KV-Ground combines high-resolution grounding with local refinement~\citep{mvp,kv_ground}.
Their next views may depend on an initial click, attention response, confidence estimate, or region policy.
An inaccurate early signal can consequently affect the visual evidence available to later refinement steps.
\ours instead selects and verifies each crop before it becomes the next observation.

Complete computer-use systems combine grounding with planning, execution, recovery, and state management.
Agent S3 applies test-time scaling, VLAA-GUI uses modular automation, HIPPO Agent adds memory, and OpenAPA targets end-to-end process automation~\citep{agent_s3,vlaa_gui,hippo_agent,openapa}.
Further configurations appear on OSWorld-Verified~\citep{osworld_leaderboard}.
These systems illustrate how grounding operates alongside broader agent components rather than as an isolated benchmark capability.
Within a complete OSWorld agent, \ours supplies the grounding capability used by these broader components.

\section{Conclusion}
\label{sec:conclusion}

In this work, we present \ours, a coarse-to-fine cropping framework for GUI grounding with general-purpose VLMs.
\ours first converts OCR and detector outputs into coordinate-aware screen references.
The VLM then selects increasingly focused crops until the target can be determined from a detailed view.
Before execution, visual verification checks the proposed target, and its local position is mapped back to the original screen.
Extensive evaluations demonstrate consistent improvements across grounding benchmarks and base models, as well as effective transfer to interactive computer-use tasks.

\bibliography{references}

\clearpage
\appendix

\section{Baseline Systems}
\label{app:baselines}

\begin{table*}[t]
\centering
\caption{ScreenSpot-v2 point-in-box accuracy (\%). Baselines are taken from the official leaderboard. Best and second-best scores in each column are bolded and underlined.}
\label{tab:screenspot_v2_comparison}
\begin{tabular}{lccccccc}
\toprule
\multirow{2}{*}{\textbf{Model / Method}} &
\multicolumn{2}{c}{\textbf{Mobile}} &
\multicolumn{2}{c}{\textbf{Desktop}} &
\multicolumn{2}{c}{\textbf{Web}} &
\multirow{2}{*}{\textbf{Avg.} }\\
\cmidrule(lr){2-3} \cmidrule(lr){4-5} \cmidrule(lr){6-7}
& \textbf{Text} & \textbf{Icon}
& \textbf{Text} & \textbf{Icon}
& \textbf{Text} & \textbf{Icon} & \\
\midrule
\multicolumn{8}{c}{General Models} \\
\midrule
GPT-4o & 26.6 & 24.2 & 24.2 & 19.3 & 12.8 & 11.8 & 20.1 \\
Qwen2-VL-7B & 52.8 & 46.4 & 47.9 & 29.3 & 29.1 & 26.1 & 39.8 \\
Qwen2.5-VL-7B-Instruct & 99.0 & 84.4 & 87.6 & 65.7 & 90.2 & 79.8 & 86.5 \\
\midrule
\multicolumn{8}{c}{Specialized GUI Models} \\
\midrule
CogAgent-18B & 69.3 & 27.0 & 75.8 & 20.7 & 74.4 & 31.5 & 52.8 \\
SeeClick-7B & 77.9 & 48.3 & 69.6 & 30.7 & 57.3 & 22.2 & 53.9 \\
UGround-7B & 84.5 & 61.6 & 85.1 & 61.4 & 84.6 & 71.9 & 76.3 \\
OS-Atlas-7B & 93.8 & 73.9 & 90.7 & 63.6 & 89.7 & 77.3 & 83.3 \\
AGUVIS-7B & 95.5 & 81.5 & 93.3 & 77.9 & 91.0 & 77.8 & 87.3 \\
Holo1.5-7B & \underline{99.2} & 91.1 & 96.7 & 88.9 & 95.4 & 86.1 & 93.3 \\
UI-Venus-72B & \textbf{99.7} & \underline{93.8} & 95.9 & 90.0 & \underline{96.2} & \underline{92.6} & \underline{95.3} \\
\midrule
\multicolumn{8}{c}{GUI Grounding Systems} \\
\midrule
GUI-ARP (7B) & 96.5 & 89.0 & 97.2 & 85.7 & 94.3 & 84.8 & 91.8 \\
UI-Zoomer (UI-Venus-7B) & 98.6 & 90.5 & \textbf{99.0} & \underline{92.9} & 95.7 & 90.6 & 94.9 \\
\midrule
\textbf{\ours (GPT-5.5)} & 98.3 & \textbf{94.8} & \underline{97.9} & \textbf{96.4} & \textbf{97.0} & \textbf{95.6} & \textbf{96.8} \\
\bottomrule
\end{tabular}
\end{table*}

The experiments compare \ours with three complementary classes of prior systems.
General-purpose VLMs test whether the proposed grounding procedure improves models that are not specialized for GUI localization.
Specialized GUI models use GUI-specific training or architectural adaptation, while GUI grounding systems add inference-time perception, refinement, or routing around a VLM.
The model names and configurations in the main tables follow the original papers, official releases, or public leaderboards.

\paragraph{Static grounding comparisons.}
ScreenSpot-Pro includes both direct model baselines and systems that refine grounding at inference time.
For the six general-purpose base models used by \ours, the matched single-shot result is obtained by asking the same backend to predict the click directly from the full screenshot.
DeepSeek-V4.1-Flash answers with a tool call and GLM-5.3-Flash with image pixels, the most accurate coordinate format of each model.
GLM-5.3-Flash receives a screenshot above its image budget already downscaled, and DeepSeek-V4.1-Flash receives every image padded to at most 16:9 and scaled to its budget, since it otherwise mixes pixel and relative coordinates on very wide or small images.
BAMI, AdaZoom-GUI, and ZoomClick follow the settings of their papers and use this call for every localization step, and the same base model also selects BAMI candidates and refines AdaZoom-GUI instructions.
UI-Zoomer follows its official code with eight sampled boxes, gating on box agreement and token confidence, and a Gaussian crop for refinement, which requires token log probabilities.
This comparison holds the VLM fixed and isolates the effect of the grounding procedure.
Results for the remaining methods are taken from their papers or the official benchmark leaderboard under the reported model configuration.
UI-Vision and MMBench-GUI-L2 follow the aligned benchmark evaluations reported by the original benchmark papers under their official protocols~\citep{ui_vision,mmbench_gui}.

\paragraph{Interactive computer-use comparisons.}
The OSWorld table uses results from the OSWorld-Verified leaderboard~\citep{osworld_leaderboard}.
Its entries include general multimodal models, specialized computer-use models, and agent systems that combine a foundation model with planning, memory, recovery, or execution modules.
Because public systems differ in backbone and agent design, the table reports their published end-to-end performance rather than treating every row as a controlled backbone comparison.
Pointer Agent provides the closest same-backbone reference because both systems use Claude Opus 4.7.
For \ours, Claude Opus 4.7 supplies high-level actions and GUI-Lens supplies the grounding procedure.
Evaluation uses a 15-step interaction limit on the Chrome, Multi-Apps, and OS domains.
Domain scores are the mean official evaluator rewards over the included tasks of each domain.

\section{Grounding Configuration and Ablations}
\label{app:grounding_ablation}

This section specifies the common grounding configuration and the controlled variants used in the component ablation.

\paragraph{Default grounding configuration.}
GUI-Lens begins from the full screenshot and constructs a sequence of increasingly focused observations.
GPT-5.5, Claude Opus 4.7, and GLM-5.3-Flash permit at most eight crop selections, while MiniMax-M3.1 permits at most five.
At each step, the prompt contains the current image, the instruction, the crop history, and up to 60 coordinate references that overlap the visible region.
The final grounding prompt permits up to 80 references.
Intermediate crops are enlarged by at most $5\times$, and the final crop by at most $8\times$, relative to their original screen resolution.
Crop boxes receive 8\% contextual padding on each side before being intersected with the screenshot boundary.
Cropping terminates when the VLM requests final grounding or reaches the maximum number of crop selections, after which the final point is predicted from the current observation.

Visual verification uses a separate stage-specific prompt with the same VLM backend.
The marked proposal is accepted or rejected according to whether it contains the instructed target.
A rejection restores the full-screen observation and starts a new crop sequence.
The retry limit is six for GPT-5.5, Claude Opus 4.7, and GLM-5.3-Flash and three for MiniMax-M3.1.
Detector and OCR outputs provide approximate references throughout this procedure; the VLM selects the crops and produces the final click.

\paragraph{Ablation protocol.}
The main paper reports leave-one-component-out results for GPT-5.5, DeepSeek-V4.1-Flash, and MiniMax-M3 on the same 300-example ScreenSpot-Pro subset.
The subset is sampled once with stratification over application category and target type using random seed 42, and its manifest is reused for every backbone and variant.
All settings other than the removed component remain unchanged.
Without Coordinate Priming, detector and OCR references are omitted from the prompt.
Without Coarse-to-Fine Cropping, the backend predicts the target from a single full-screen observation.
Without Visual Verification, crop and click proposals are applied without the verification call.
Each change in the table is measured against the full configuration evaluated on the identical subset of 300 examples.

\section{Per-Benchmark Detailed Results}
\label{app:per_domain}

Table~\ref{tab:screenspot_v2_comparison} reports the detailed ScreenSpot-v2 results omitted from the main paper.
The benchmark separates Mobile, Desktop, and Web interfaces into Text and Icon targets.

\paragraph{Platform breakdown.}
GUI-Lens reaches 98.3/94.8 on Mobile, 97.9/96.4 on Desktop, and 97.0/95.6 on Web for Text/Icon targets, respectively.
The resulting 96.8 average exceeds the strongest reported baseline by 1.5 points.
The gains therefore extend across all three interface types.

\paragraph{Target-type breakdown.}
GUI-Lens obtains the highest Icon accuracy on all three platforms.
Relative to the strongest baseline in each Icon column, the gains are 1.0 points on Mobile, 3.5 points on Desktop, and 3.0 points on Web.
This pattern shows the benefit of focused observations for visually defined targets such as icons without text labels.

\section{Implementation Details}
\label{app:implementation}

\paragraph{Screen perception.}
GPA-GUI-Detector~\citep{gpa-gui-detector} extracts visual components with a confidence threshold of 0.1 and an NMS IoU threshold of 0.3.
EasyOCR~\citep{baek2019craft,shi2017crnn} recognizes English and Simplified Chinese text.
Both modules run before the first VLM call.
Each retained reference records its label, bounding box, center, confidence, and source type.
These values are serialized as approximate spatial references and do not determine the target independently of the image.

\paragraph{Endpoints and image budgets.}
Each base model is called through one endpoint with that endpoint's default image settings, and Table~\ref{tab:endpoints} lists the endpoint and image budget of every base model in the paper.
An endpoint encodes an image as visual tokens that each cover a fixed pixel patch and downscales a screenshot whose token count would exceed its limit, so the budget is the token limit times the patch area.
We find the limit by sending square test images of increasing size until the image token count stops growing.
Claude Opus 4.7 runs through the Claude Code subscription, which resizes every image whose long edge exceeds 2{,}000 pixels to a long edge of 2{,}000 pixels while keeping its aspect ratio.

\begin{table}[t]
\centering
\small
\setlength{\tabcolsep}{4pt}
\caption{Endpoint and image budget of each base model, with the base models of Table~\ref{tab:screenspot_pro_comparison} on top and those of Section~\ref{sec:exp_backbone} at the bottom. Tokens is the largest number of visual tokens per image, Patch the pixels covered by one token, and MP the largest image in megapixels that the endpoint sends without downscaling. Claude Opus 4.7 is limited to a long edge of 2{,}000 pixels, which is 2.25 megapixels for a 16:9 screenshot.}
\label{tab:endpoints}
\begin{tabular}{@{}llrcr@{}}
\toprule
\textbf{Base model} & \textbf{Endpoint} & \textbf{Tokens} & \textbf{Patch} & \textbf{MP} \\
\midrule
GPT-5.5 & Codex & 10{,}000 & $32\times32$ & 10.24 \\
Claude Opus 4.7 & Claude Code & -- & -- & 2.25 \\
DeepSeek-V4.1-Flash & DeepSeek API & 1{,}024 & $42\times42$ & 1.81 \\
MiniMax-M3 & MiniMax API & 5{,}186 & $28\times28$ & 4.07 \\
GLM-5.3-Flash & Ark, OpenCode Go & 7{,}923 & $28\times28$ & 6.21 \\
\midrule
GPT-6.1-Sol & Codex & 30{,}000 & $32\times32$ & 30.72 \\
GPT-6-Luna & Codex & 30{,}000 & $32\times32$ & 30.72 \\
MiniMax-M3 & OpenCode Go & 5{,}387 & $28\times28$ & 4.22 \\
MiniMax-M3.1 & OpenCode Go & 5{,}184 & $28\times28$ & 4.06 \\
GLM-5.3-Flash & OpenCode Go & 7{,}923 & $28\times28$ & 6.21 \\
DeepSeek-V4.1 & DeepSeek API & 1{,}024 & $42\times42$ & 1.81 \\
Llama 4 Maverick & OpenRouter & 2{,}448 & $28\times28$ & 1.81 \\
Gemma 3 27B & OpenRouter & 256 & $56\times56$ & 0.80 \\
\bottomrule
\end{tabular}
\end{table}

\paragraph{Prompt and response protocol.}
Crop proposal, visual verification, and final grounding use separate calls to the same backend with prompts specialized for their respective decisions.
The evaluated backends are GPT-5.5, Claude Opus 4.7, DeepSeek-V4.1-Flash, MiniMax-M3, and GLM-5.3-Flash.
A crop-proposal call receives the instruction, current observation, visible coordinate references, and crop history, then returns either a crop box or a request for final grounding.
The crop box is represented as \texttt{[x1, y1, x2, y2]} in the displayed image coordinates.
A verification call receives the same instruction and an image marked with the proposed crop or click, then returns \texttt{accept} or \texttt{reject}.
The final-grounding call returns either a coordinate-reference identifier or an explicit point in the current image.
All responses use a constrained JSON schema that is validated before execution.

\paragraph{Coordinate transformation.}
Suppose the current crop occupies $[x_0,x_1)\times[y_0,y_1)$ in the original screenshot and is displayed to the VLM at width $W_r$ and height $H_r$.
A displayed point $(u,v)$ is mapped back to the original screen by
\begin{equation}
    \Phi_r(u,v)=
    \left(x_0+\frac{x_1-x_0}{W_r}u,\;
          y_0+\frac{y_1-y_0}{H_r}v\right).
\label{eq:app_coordinate_map}
\end{equation}
The same affine transformation is applied to both corners of a proposed crop box.
Mapped regions are intersected with the screenshot boundary before cropping, and the accepted final point is returned in original screenshot coordinates.

\paragraph{Evaluation protocol.}
Static benchmarks use their official point-in-box evaluation: a sample is correct when the predicted point lies inside the annotated target box.
OSWorld uses the official task evaluators for end-to-end completion.
Each reported number is computed from one evaluation of every included sample or task under the stated configuration; the paper does not average repeated API runs.

\paragraph{Reproducibility assets.}
The code supplement contains the grounding pipeline, benchmark adapters, configuration files, and instructions required to reproduce the reported evaluation procedure, and excludes credentials and the cached model responses of our evaluation runs.

\section{Base Model Analysis Details}
\label{app:backbone}

This section gives the protocol behind Section~\ref{sec:exp_backbone}.

\paragraph{Subset.}
The probes and the \ours runs use 100 ScreenSpot-Pro examples drawn with a fixed seed and stratified by application group and target type (text or icon).
Among 20{,}000 draws, we keep the one that best matches the full benchmark in the distribution of relative target area and screenshot resolution and in the accuracy of an earlier \ours run with DeepSeek-V4.1 (71.0\% vs.\ 71.4\%).

\paragraph{Base models and endpoints.}
Each base model in Table~\ref{tab:backbone_probes} is one model at one reasoning effort through one endpoint, and all of its probes and its \ours run share that setting.
DeepSeek-V4.1 uses the official API; GLM-5.3-Flash, MiniMax-M3.1, and MiniMax-M3 use the OpenCode Go gateway; Llama~4 Maverick and Gemma~3 27B use OpenRouter; and the two GPT base models use the Codex endpoint.
Each endpoint downscales a screenshot that exceeds its image budget (Table~\ref{tab:endpoints}), and only the GPT base models receive every screenshot at full resolution.

\paragraph{Probes.}
\textit{Direct click} ($D$) asks for the click point on the full screenshot in four coordinate formats: pixels of the image, relative coordinates in $[0,1000]$, fractions in $[0,1]$, and a tool call with a \texttt{coordinate} argument.
$D$ takes the format with the highest accuracy for that base model (Table~\ref{tab:backbone_formats}), which also covers GLM-4.6V, GLM-5.3-FlashX, Gemini 4 Argon, Grok 4.7, Kimi K2.5, MiMo-V2.6-Pro, Mistral Small 3.2, Step-3.7-Flash, and MiniMax-VL-01~\citep{glm46v,glm53_flash,kimi_k25,mimo_v26,minimax_vl01}.
\textit{Coarse localization} ($C$) reuses the click of that format: a window of half the screen width and height is centered on the click, shifted to stay inside the screen, and $C=1$ if it contains the whole target box.
\textit{Knowledge} ($K$) shows a quarter-screen crop around the target, upscaled to about 2 megapixels, together with up to eight numbered tiles, one for the target and the others for nearby detected elements; the base model only names the tile number, so $K$ measures recognition of the target alone.
\textit{Reasoning} ($R$) is $\log_{10}(1+t)$, where $t$ is the endpoint-reported mean of reasoning tokens per call in the knowledge probe.

\paragraph{Fit.}
Eq.~\eqref{eq:gain} is fitted by least squares without an intercept on the per-example gain $g-D\in\{-1,0,1\}$, where $g$ indicates that \ours hits the target, over all 800 model--example pairs.
The resolution $b$ is $\log(B/0.80)/\log(30.72/0.80)$ for an image budget of $B$ megapixels, which gives 0.00 for Gemma~3 27B, 0.22 for DeepSeek-V4.1 and Llama~4 Maverick, 0.45 and 0.46 for the two MiniMax base models, 0.56 for GLM-5.3-Flash, and 1.00 for the GPT base models.
For the held-out prediction of each base model, the coefficients are refitted without its 100 examples; across the eight refits they range over $\alpha\in[0.27,0.33]$, $\gamma\in[0.16,0.17]$, and $\rho\in[0.16,0.18]$.
Table~\ref{tab:backbone_probes} lists the probe scores and the held-out predictions.
Removing one term at a time, the held-out mean absolute error rises from 3.3 points to 8.5 without the zoom term, 12.5 without reasoning, and 7.3 without resolution; using $C-D$ without $K$ in the zoom term gives 3.5.

\paragraph{Target size.}
For Figure~\ref{fig:size_gain}, the side of a target in vision patches is $\sqrt{wh}\cdot\min\!\big(1,\sqrt{B/(WH)}\big)/p$, where $w\times h$ is the target box, $W\times H$ the screenshot, $B$ the endpoint's image budget in pixels, and $p$ its patch size (42 pixels for DeepSeek-V4.1, 32 for the GPT base models, 56 for Gemma~3 27B, and 28 for the others).
The four bins hold 169, 244, 257, and 130 of the 800 model--example pairs.

\paragraph{Errors of GPT-6.1-Sol.}
GPT-6.1-Sol solves 96 of the 100 examples by direct click and 95 with \ours.
\ours recovers two small targets the direct click misses, including ``Chip planner'' in Section~\ref{sec:exp_backbone}, and loses three examples that the direct click solves, in all three runs.
In these three, the final grounding step decides again on the enlarged crop without the reasoning of earlier rounds and picks a visually similar element, such as the original file instead of its copy.
Passing the earlier rounds to the final step fixes one of the three but loses another example, leaving accuracy at 95.

\begin{table}[t]
\centering
\small
\setlength{\tabcolsep}{3pt}
\caption{Probe scores (\%) and gain of \ours over the direct click (points) for the eight base models of Section~\ref{sec:exp_backbone}. Tok.\ is the mean number of reasoning tokens per call in the knowledge probe, $b$ the image resolution of the endpoint, Pred.\ the gain predicted by Eq.~\eqref{eq:gain} with the base model held out, and Meas.\ the measured gain.}
\label{tab:backbone_probes}
\resizebox{\linewidth}{!}{%
\begin{tabular}{@{}llrrrrrrrrr@{}}
\toprule
\textbf{Base model} & \textbf{Effort} & $\boldsymbol{D}$ & $\boldsymbol{C}$ & $\boldsymbol{K}$ & \textbf{Tok.} & $\boldsymbol{R}$ & $\boldsymbol{b}$ & \textbf{Pred.} & \textbf{Meas.} & \textbf{\ours} \\
\midrule
Gemma 3 27B & off & 1 & 36 & 16 & 0 & 0.00 & 0.00 & $+18$ & $+19$ & 20 \\
Llama 4 Maverick & off & 6 & 80 & 64 & 0 & 0.00 & 0.22 & $+29$ & $+27$ & 33 \\
GLM-5.3-Flash & low & 19 & 91 & 85 & 33 & 1.53 & 0.56 & $+43$ & $+44$ & 63 \\
DeepSeek-V4.1 & low & 22 & 78 & 78 & 521 & 2.72 & 0.22 & $+58$ & $+58$ & 80 \\
MiniMax-M3.1 & low & 39 & 89 & 90 & 196 & 2.29 & 0.45 & $+43$ & $+38$ & 77 \\
MiniMax-M3 & low & 46 & 92 & 94 & 748 & 2.87 & 0.46 & $+45$ & $+37$ & 83 \\
GPT-6-Luna & medium & 82 & 94 & 88 & 28 & 1.46 & 1.00 & $+7$ & $-2$ & 80 \\
GPT-6.1-Sol & medium & 96 & 98 & 99 & 3 & 0.60 & 1.00 & $+1$ & $-1$ & 95 \\
\bottomrule
\end{tabular}%
}
\end{table}

\begin{table}[t]
\centering
\small
\setlength{\tabcolsep}{4pt}
\caption{Direct-click accuracy (\%) on the 100-example subset in each coordinate format. Asterisks mark the base models of Section~\ref{sec:exp_backbone}, and bold marks the format used for $D$. The same base model can differ by 30--50 points across formats.}
\label{tab:backbone_formats}
\begin{tabular}{@{}lrrrr@{}}
\toprule
\textbf{Base model} & \textbf{Pixel} & \textbf{0--1000} & \textbf{0--1} & \textbf{Tool} \\
\midrule
GPT-6.1-Sol$^\ast$ & \textbf{96} & 94 & 94 & 91 \\
GPT-6-Luna$^\ast$ & \textbf{82} & 78 & 74 & 3 \\
Kimi K2.5 & 1 & 32 & 56 & 3 \\
Step-3.7-Flash & 0 & 51 & 0 & 36 \\
Gemini 4 Argon & 13 & 9 & 38 & 49 \\
MiniMax-M3$^\ast$ & 31 & \textbf{46} & 39 & 17 \\
GLM-4.6V & 1 & 42 & 43 & 34 \\
Grok 4.7 & 41 & 32 & 38 & 14 \\
MiniMax-M3.1$^\ast$ & 35 & \textbf{39} & 38 & 5 \\
MiMo-V2.6-Pro & 0 & 11 & 38 & 5 \\
DeepSeek-V4.1$^\ast$ & 0 & 18 & 18 & \textbf{22} \\
GLM-5.3-FlashX & 19 & 7 & 11 & 0 \\
GLM-5.3-Flash$^\ast$ & \textbf{19} & 3 & 8 & 0 \\
Llama 4 Maverick$^\ast$ & 1 & \textbf{6} & 5 & 4 \\
Mistral Small 3.2 & 0 & 2 & 0 & 0 \\
MiniMax-VL-01 & 1 & 2 & 2 & 1 \\
Gemma 3 27B$^\ast$ & 0 & 0 & 0 & \textbf{1} \\
\bottomrule
\end{tabular}
\end{table}

\end{document}